\documentclass{article}
\PassOptionsToPackage{comma,numbers,sort&compress}{natbib}

\usepackage{arxiv}

\usepackage[utf8]{inputenc} 
\usepackage[T1]{fontenc}    
\usepackage{hyperref}
\usepackage{url}            
\usepackage{booktabs}       
\usepackage{amsfonts}       
\usepackage{nicefrac}       
\usepackage{microtype}      
\usepackage{color}
\usepackage{adjustbox}
\usepackage[symbol]{footmisc}

\usepackage{mleftright,xparse}
\usepackage{amsmath}
\usepackage{amssymb}
\usepackage{indentfirst}
\usepackage{mathtools}
\usepackage{xparse}
\usepackage{mathrsfs}

\usepackage{graphicx}
\graphicspath{ {Images/} }

\NewDocumentCommand\xDeclarePairedDelimiter{mmm}
 {%
  \NewDocumentCommand#1{som}{%
   \IfNoValueTF{##2}
    {\IfBooleanTF{##1}{#2##3#3}{\mleft#2##3\mright#3}}
    {\mathopen{##2#2}##3\mathclose{##2#3}}%
  }%
 }
\xDeclarePairedDelimiter{\brac}{(}{)}
\xDeclarePairedDelimiter{\sbrac}{[}{]}
\xDeclarePairedDelimiter{\fbrac}{\{}{\}}
\xDeclarePairedDelimiter{\norm}{\lVert}{\rVert}
\xDeclarePairedDelimiter{\abs}{\lvert}{\rvert}
\DeclarePairedDelimiterX{\inp}[2]{\langle}{\rangle}{#1, #2}

\usepackage[natbib, maxcitenames=3, style=numeric, backend=biber, url=false, isbn=false, doi=false, sorting=none]{biblatex}
\bibliography{./reference}

\hypersetup{
    colorlinks,
    citecolor=[rgb]{.282, .449, .718},
    linkcolor=[rgb]{.282, .449, .718},
}

\let\oldcitep=\citep
\renewcommand\citep[1]{\hypersetup{linkcolor=[rgb]{.2, .2, .4}}\hyperlink{#1}{\oldcitep{#1}}}

\title{Irregular Convolutional Auto-Encoder on Point Clouds}

\author{%
  Zhang Yuhui \\
  Tokyo Institute of Technology \\
  \texttt{zhang.y.av@m.titech.ac.jp} \\
  \And
  Greg Gutmann \\
  Tokyo Institute of Technology \\
  \texttt{gutmann.g.aa@m.titech.ac.jp} \\
  \And
  Konagaya Akihiko \\
  Tokyo Institute of Technology \\
  \texttt{konagaya.a.aa@m.titech.ac.jp} \\
}

\begin{document}

\maketitle

\begin{abstract}
    We proposed a novel graph convolutional neural network that could construct a coarse, sparse latent point cloud from a dense, raw point cloud. With a novel non-isotropic convolution operation defined on irregular geometries, the model then can reconstruct the original point cloud from this latent cloud with fine details. Furthermore, we proposed that it is even possible to perform particle simulation using the latent cloud encoded from some simulated particle cloud (e.g. fluids), to accelerate the particle simulation process. Our model has been tested on ShapeNetCore dataset for Auto-Encoding with a limited latent dimension and tested on a synthesis dataset for fluids simulation. We also compare the model with other state-of-the-art models, and several visualizations were done to intuitively understand the model.
\end{abstract}


\section{Introduction}

A huge amount of real-life objects could be efficiently represented as particles, or point clouds. 
General 3D Objects, radar scans, fluids, or even atoms and molecules. Point clouds are an efficient and clean representation of real-world objects, as they directly define a distribution using points as samples. 
There are also other methods to represent 3D objects, two of the most common usage is polygon meshes and volumetric data arrays. 
Polygon (typically triangular) meshes are restricted to preserve a topology between vertices, and usually they can only represent the surface of a given object. 
Volumetric data arrays usually have a regular lattice structure by definition, which is not accurate since objects typically does not follow a regular lattice structure and some noise was introduced when quantizing them to those regular lattices. 

Point clouds, however, are the most simple yet accurate way to represent objects among those three representations. 
Firstly, it is simple since it does not need to have topological relationships between points. 
Secondly, it is accurate since it does not need to quantize original objects into some grid bins like volumetric representations. 
Last but not the least, point clouds are very flexible as they could easily represent anything, and they are the easiest to obtain, usually could be directly obtained from data sensors (e.g. LiDAR, satellites) without any pre-processing.
Thus, it is very promising and valuable to directly operate on point clouds.
However, those point clouds often come with a massive amount of data, introduced a big challenge while processing such particle clouds. 
The huge amount of data is also a problem for other representation methods as well. 

The goal of representation learning is to find powerful, rich and meaningful representations of some complex high-dimensional objects. 
As inspired by human's ability of information summarization without a teacher, learning those representations are often preferred in an unsupervised manner.
Auto-encoding tasks then become a fundamental task in unsupervised representation learning literature. 
The origin of learning algorithms with auto-encoders was widely-agreed as the work done by \citep{AEOrigin}. 
Auto-encoding typically trains a model by first encoding input data to a smaller latent space (compared to original input space), then reconstructs it as fine as possible. 
The latent code is often expected as a well-behaved representation, as lots of experiments show that interpolating them provides meaningful results.

Recently, deep auto-encoders have been proven to have a strong capability in finding latent representations of data in an unsupervised manner, such as images \citep{VQVAE, VQVAE2, CAE_ICANN}, 3D volumes \citep{LatentPhys, DeepFluids}, polygon meshes \citep{tanMeshAE, meshVAE}, videos \citep{STAE_Video, STAE_Video_abnomral}, texts \citep{seq2seq}, as well as point clouds \citep{fold, rawgenpc}. 
In this work, we aim to learn light-weight and information-rich representations of point clouds by a graph-based deep auto-encoder, which could benefit in many different ways. 
For example, we could achieve a better understanding of different shapes by those deep representations, or we could process those deep representations directly rather than the raw point cloud to save computational budget. 
In the latter case, particularly in this work, we are interested in performing latent-space physics simulation with those deep representations. 
By doing so, we could achieve similar results compared with raw simulation on particles, with a huge speed boost as our representations are light-weight.
Typically, raw particle clouds contain thousands to millions of particles, while our representation only contains tens to hundreds of information-rich particles. 
In this paper, we will propose a novel point cloud auto-encoder architecture, based on a novel convolution operation defined over irregular point clusters. 

\subsection{Our contributions}
We summarize our contributions in this paper as follows:

\begin{itemize}
    \item We proposed a novel convolution layer for point clouds. Our convolution layer works without any discretization so that it could directly operate on accurate point clouds. Furthermore, we pointed out and proposed a simple solution to a possible problem that only occurs when the input data distribution is irregular.
    \item We combined our novel convolution layer with farthest-sampling as pooling methods, which gave us a graph-based deep convolutional neural network that learns representations for point clouds. We extended our encoder by a local-oriented, conditional generator as a decoder, and trained it in an auto-encoding manner. Our novel deep auto-encoder achieved state-of-the-art performance on such tasks. We also provided a detailed analysis of it.
    \item Besides, we trained a simple yet effective learn-able dynamics estimator to learn the particle dynamics under the encoded latent space. We used fluids (which are based on Navier-stokes equations) as our dataset for training and evaluation. Our model successfully learned the dynamics under the latent space, shows that our auto-encoder has the ability to produce senseful representations.
\end{itemize}

\subsection{Notations}

We now state some notations that will be used throughout this paper. 
We denote $p_i$ for the position of $i$-th point inside the point cloud, 
$\mathcal{N}\brac{i}$ as the set of indices of points inside the neighborhood of $i$-th point, 
$\mathbf{f}^{\brac{\ell}}$ as the output feature map of layer $\ell$ (defined as a vector-valued function over $\mathbb{R}^{3}$), 
$\mathbf{f}^{\brac{\ell}}_{i}$ as the sampled value of the feature map function on the position of $i$-th point (which is more widely used in this work), 
$P\brac{X}$ the underlying point distribution of the input point cloud, 
with an intractable probability density function (PDF) $\mu \brac{x}$ and we sometimes also denote it as $\norm{\mathbf{f}^{\brac{0}}}$ 
\footnote{For consistency, we still write $\mathbf{f}^{\brac{0}}$ as vector form even it is real-valued and the norm operator $\norm{\cdot}$ is unnecessary.} 
since it is the desired input to our model.

\section{Related works}

\paragraph{Auto-encoders and representation learning} Variational auto-encoders (VAEs) \citep{VAE} provided a powerful extension to auto-encoders by constructing a variational lower bound to maximize the likelihood, as well as the ability to directly sample from learned data distributions. 
\citep{CAE_ICANN} used a stacked convolutional auto-encoder on images, while \citep{fold, rawgenpc, ShapeVAE, meshVAE, tanMeshAE} operates on 3D shapes, and \citep{advgae} operates on topological graphs. 
\citep{VQVAE, VQVAE2} achieved state-of-the-art results in image synthesis by an auto-regressive variational auto-encoder, showing the power of an auto-encoder and an auto-encoding learning task. 
\citep{RotPredictAE} trained an auto-encoder to predict how the rotation of a 3D object looks in a 2D image, resulting in useful representations.
\citep{AdaIN} found that the moments (i.e. mean and variance) of the activation of intermediate layers (i.e. intermediate feature maps) are often corresponded to the style of an image, while the "relative position" of features contains semantical contents, and have achieved arbitrary style transfer between images by simple transformations.

\paragraph{Generative models} Meanwhile, generative models are also attractive for their abilities to estimate generate samples from very complex unknown empirical distributions, e.g. image and video synthesis. 
As we treat point clouds as point distributions in a 3-D space, we are particularly interested in generating such distributions, conditioned on learned latent representations. 

Generative adversarial networks \citep{GAN} provide a powerful framework with mind-blowing results in diverse generative tasks, such as image synthesis \citep{CGAN, SAGAN, WGAN_GP, BigGAN, StyleGAN, CycleGAN, NVGrowGAN}, and image-to-image translation \citep{CycleGAN, CycleGAN_HideInfo, pix2pix}. 
To our interest, such tasks often focused on a conditional distribution where the conditioning methods for the generator could vary across different works \citep{CGAN, StyleGAN, AdaIN}. 

On the other hand, the discriminator acts like a statistical distance between fake generated distribution and real distribution \citep{WGAN_pre}, in particular, they could be KL divergence estimators in vanilla GAN configuration \citep{GAN}, Wasserstein distance \citep{WGAN, WGAN_GP}, Total Variation \citep{EBGAN}, Maximum Mean Discrepancy \citep{GMMN, cGMMN, MMD15, MMD17}, f-divergence \citep{fGAN} and so on. 

As point clouds are low-dimensional data distributions so that distance estimation could be relatively easy and accurate, a recent work \citep{PCGAN} adds a non-parametric upper-bound approximation for Wasserstein distance \citep{EMDUB} on top of Wasserstein GANs \citep{WGAN, WGAN_GP} for point cloud generation. 
Sinkhorn distance \citep{Sinkhorn} are also good estimators and solvers for Wasserstein distance and optimal transport, while it introduced an entropical constraint to the transport plan matrix in the original optimal transport problem, enabled it to use a simple iterative method to solve this constrained problem as an accurate estimator to the original problem.

\paragraph{Graph convolutional networks (GCNs)} Graphs are very powerful mathematical representations for relationships. 
Especially, in point clouds, we could define such relationships by the spatial relationships of points, namely a k-Nearest Neighbor Graph (k-NNG). 
Thus we could focus on some local neighbors of a point to extract features hierarchically. 
Many machine learning methods were developed for such graph representations. 

Spectral methods were applied as the convolution in the spatial domain represented as multiplication in the spectral (Fourier) domain.
This could be well-approximated via Chebyshev polynomials up to a given order as suggested in \citep{graphwavelets}, and was used to build GCNs in \citep{chebynet}. 
A first-order approximation was given by \citep{semiGCN}, hence the convolution was defined over the 1-neighbors (i.e. 1 step away) of a given node in the graph. 
Those convolution operations are isotropic (i.e. weights are the same for different neighbors with the same order), as they are all equals to $W$ in equation \ref{Eq:specG}.
In contrast, conventional convolution kernels often have different weights for different input nodes.
To tackle this problem, attention mechanism has been used as it could let the model focus on different parts in the output for different neighbors \citep{gattention}. 
The layer-wise convolution operations are typically defined as \citep{semiGCN}: 
\begin{equation} \label{Eq:specG}
    X_{\ell + 1} = \sigma \brac{ \hat{D}^{-\frac{1}{2}} \hat{A} \hat{D}^{-\frac{1}{2}} X_\ell W_\ell }
\end{equation}
where $W_\ell$ is the weight matrix, $\hat{A} = A + I$ is the graph adjacency matrix $A$ with self-loops added, $\hat{D}$ is the diagonal node degree matrix, used to normalize $\hat{A}$. 
Such spectral methods often require the graph topology to be fixed along training, therefore they do not fit in our conditions.

Other methods follow a message-passing scheme between vertices \citep{IN, HRN, MPNN, deepmind}. 
In those works, convolution is often defined as a combination of functions over graph edges. 
For example, in Interaction Networks \citep{IN}, layer-wise convolution operations for a single vertex are defined as equation \ref{eq:IN} in section \ref{Chpt:IN}.

Besides convolution, pooling operations are also important since it could reduce computational cost and let the convolution operation operates on a larger receptive field. 
Non-parametric methods, such as binary tree indexing \citep{chebynet} and graph clustering \citep{gpoolCluster, DQI}, were used to coarsen the graph, while some works used a learnable pooling operation \citep{gpoolMat, gunet} for graph coarsening. 
As suggested in \citep{PointNet++, PCSpectralGConv}, farthest sampling could be very efficient in pooling for graphs with spatial information (e.g. k-NNG of point clouds) as they could coarsen the graph uniformly in space.

\paragraph{Geometric convolution operations} As 3D geometries (e.g. point clouds, polygon meshes) have spatial features besides topological connections, many works are done in this field focused on spatial-sensitive convolution operations. 
Lots of approach aims to use regular 3d convolution kernels as they firstly pre-process point cloud or geometric data into voxels. 
Per-point networks followed by a collector network is also a common approach \citep{PointNet, PointNet++, deepsets}. 
Besides, \citep{SplatNet} developed a similar approach that first map irregular point cloud into a regular lattice, and map them back to point cloud after regular convolution. 
\citep{sphereconv} used spherical (angular and radial) bins to discretize the original input point cloud to regular convolution operations with different weight matrices defined on different bins.
\citep{SpecConv} constructs a convolution method based on spectral methods. 
\citep{xConv} used a permutation matrix to do the convolution on a local neighborhood while keeping the permutation invariance and also enables the network to efficiently collect spatial features. 
\citep{kernelConv} presents a novel method that parameterizes each convolution kernel to small-sized point clouds (e.g. 4 points), and their convolution operation works as the computation of kernel correlation between the local point cloud and convolution kernel. 
However, none of those methods have considered the density of input point cloud distribution, which could be crucial for further development and deep models.

\paragraph{Deep learning with point clouds and 3D geometries} Traditional point cloud processing algorithms were often based on feature histograms. 
Recently, many learning-based methods were studied \citep{PointNet, PointNet++, fold, SegCloud, deepsets}. 
Several approaches used per-point feature extractor multi-layer perceptrons (MLPs) along with a global combination operation \citep{PointNet, deepsets}, which were often chosen as max-pooling or summation. 
PointNet++ \citep{PointNet++} extended \citep{PointNet} by hierarchially stacking PointNet \citep{PointNet} with farthest sampling as pooling operation. 

\citep{GeneralGraphPC} proposed a general graph network framework for point cloud machine learning methods rather than some novel convolution operations or point out important problems as we did in our paper.
\citep{PCPool} mainly focused on pooling layers occurred in point cloud processing networks, as they tried to maintain the overall structure of the original point cloud while pooling.
\citep{SuperPt} defined super-points in the similar manner of super-pixels in image processing field, using some simple geometrical structures.
\citep{SuperVoxel} was an extension of voxel-based methods, and they designed their method to preserve
as much information as possible during discretization from raw points to voxels.
In contrast, our purposed method was different from above approaches, as our model operates directly on raw points without discretization, and we defined a novel convolution operation and we pointed out an important problem in irregular data processing along with two different approaches purposed by us in order to solve that problem.

Meanwhile, there are also many works aims for generating and auto-encoding point clouds and 3D shapes \citep{FHQ, fold, rawgenpc, multiviewpc, PCGAN, PC_PCAGAN, pix2mesh, meshVAE, tanMeshAE, ShapeVAE}. 
\citep{FHQ, rawgenpc} used a deterministic approach to generate point clouds as a whole object, while \citep{fold, PCGAN, PC_PCAGAN} used conditional generators to learn a transformation from a trivial distribution to the desired point cloud.

While generating point clouds, statistical distances are widely used as the objective function for generative models. 
Chamfer pseudo-distance is widely used \citep{FHQ, fold, rawgenpc}. 
However \citep{rawgenpc} found that Chamfer pseudo-distance is blind to point densities and suggests Wasserstein distance estimators. 
Those estimators were also widely used \citep{FHQ, rawgenpc, PCGAN} in this field. 

\paragraph{Physics simulation with deep learning} Recently there are some works aimed to perform physics and quantum motion dynamics with learning methods \citep{IN, HRN, DQI, MPNN}, where fluids simulation are commonly used as a showcase for those works. 
Some of them target on 3-D spaces directly, aimed for a differentiable simulator \citep{IN, HRN, DQI}, while others targets on an encoded latent space representation \citep{LatentPhys, DeepFluids, MLMD_E1} for fast simulations. 
Learnable methods are nearly the only choice for latent dynamics since they are typically intractable. 

Almost all of latent space dynamics estimators targets to a Eulerian point of view \citep{DeepFluids, MLMD_E1, LatentPhys}, where they use vector and scalar fields as a representation instead of particles or points. 
To the best of our knowledge, few prior works were done as a latent dynamics estimator for massive amount of particles under a Lagrangian point of view, possibly caused by the obstacles while dealing with irregular points data. 
Lagrangian methods could have arbitrary large simulation space, and more efficient as the computational cost is not sensitive to the size of simulation space, compared to Eulerian methods.

\section{The proposed point cloud Auto-Encoder} \label{Chpt:AE}

\begin{figure}[t]
    \makebox[\textwidth][c]{\includegraphics[width=1.0\textwidth]{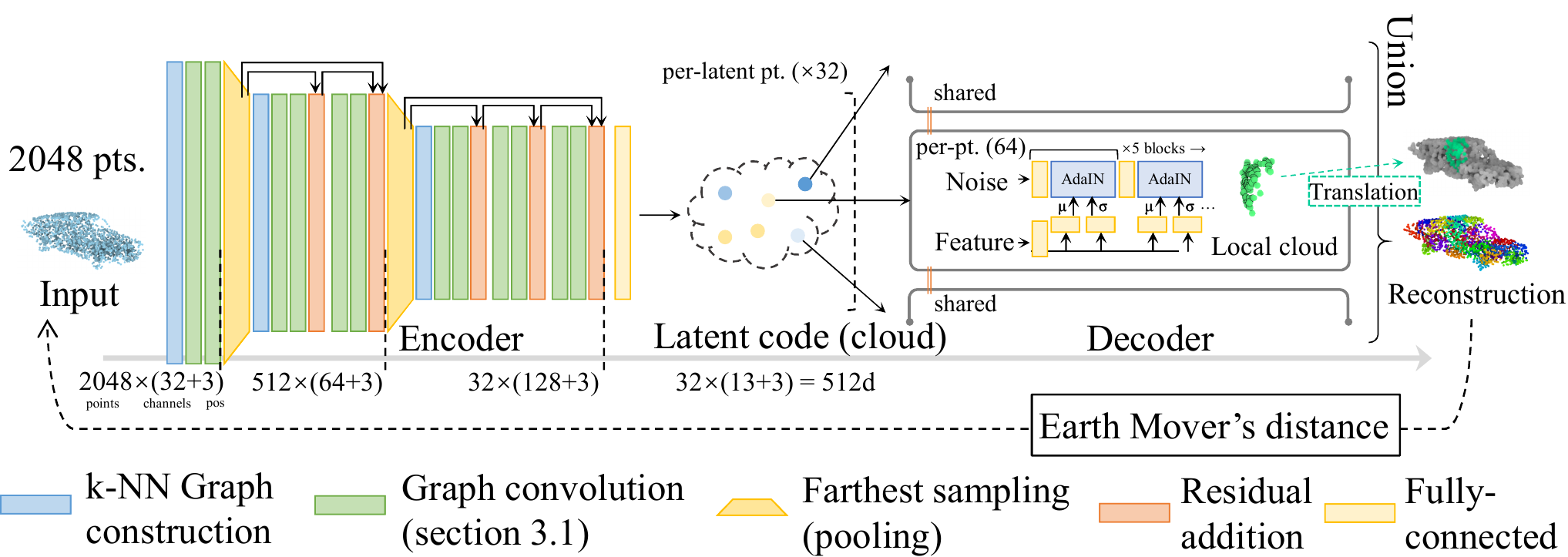}}%
    \caption{Over-all architecture of our model. As we described in section \protect\ref{Chpt:AE}, our encoder was built with hierarchical blocks, where the size of each block $N \times \brac{C + 3}$ are as $N$ points with $C$ channels and 3-d positions. Our decoder was composed of local conditional generators, as we highlighted specific local patches in the figure with different colors. All the visualizations about point clouds in this work are done with CloudCompare \protect\citep{CloudCompare}.}
    \label{fig:overallArch}
\end{figure}

Our model is built up by k-Nearest Neighbor Graphs, followed by novel convolutional layers combined with farthest-sampling pooling layers \citep{PointNet++} which are connected iteratively with residual connections, ends up with a generator conditioned on latent features by Adaptive Instance Normalization (AdaIN) \citep{AdaIN} as the decoder, shown by figure \ref{fig:overallArch}. 
Section \ref{Chpt:conv} and section \ref{Chpt:convExt} states our convolution operation, while section \ref{Chpt:AEEnc} introduces the encoder architecture and \ref{Chpt:AEDec} discussed about the decoder. 
Finally, \ref{Chpt:AETrain} shows different choices for the loss function and the training process.

\subsection{Convolution} \label{Chpt:conv}
Typically, convolution over a regular lattice (e.g. word embeddings, pixels and voxels) are defined as (in 1-D):
\begin{equation} \label{Eq:discrete_1dconv}
    f_{i}^{\brac{\ell + 1}} = \sum_{j = - k}^{k}{w_{j} f_{i + j}}
\end{equation}
This definition and parameterization are highly relied on a regular grid layout of data points, which does not hold in point cloud processing scenarios. 
Recall the definition of convolution:
\begin{equation} \label{Eq:conv}
    \brac{w \ast f}\brac{x} = \int_{-\infty}^{\infty}{w\brac{\tau} f\brac{x - \tau} d \tau}
\end{equation}
It is clear that equation \ref{Eq:discrete_1dconv} is a discrete and truncated version of equation \ref{Eq:conv}, as the weights of conventional convolution operations are discrete functions of local positions. 
Such parameterization was only possible with a regular data lattice. 
Consider point distribution as the input, we could then rewrite equation \ref{Eq:conv} for point distributions as $f^{\brac{\ell + 1}}\brac{x} = \int_{\mathcal{N}\brac{0}}{w\brac{\tau} f^{\brac{\ell}}\brac{x + \tau} d \tau}$, where $f^{\brac{0}}\brac{x} = \mu \brac{x}$ is the PDF of input point distribution. 
Note that here we truncated the convolutional kernel within a neighborhood $\mathcal{N}\brac{x}$. 
Feature maps of each layer are then given by $f^{\brac{1}}, f^{\brac{2}}, \dots, f^{\brac{L}}$, and they are not necessarily to be a valid probability density function. 
By extending above to vector forms, we obtained:
\begin{equation} \label{Eq:cconv_primal}
    \mathbf{f}_{i}^{\brac{\ell + 1}} = \int_{\mathcal{N}\brac{\mathbf{0}}}{\mathbf{W}^{\brac{\ell}}\brac{\tau}^T \mathbf{f}^{\brac{\ell}}\brac{p_i + \tau} d \tau} \approx \sum_{j \in \mathcal{N}\brac{i}}{\mathbf{W}_\theta^{\brac{\ell}} \brac{p_j - p_i}^T \frac{\mathbf{f}_{j}^{\brac{\ell}}}{\mu \brac{p_j}} }
\end{equation}
Where $\mathbf{W}^{\brac{\ell}}_\theta \brac{\cdot} \in \mathbb{R}^{3} \rightarrow \mathbb{R}^{c^\ell \times c^{\ell + 1}}$ is a parametric function as the convolutional kernel weights with learnable parameters $\theta$, and $\mathcal{N}\brac{i}$ is the 1-neighborhood of point $i$ in the k-NNG of input point cloud. 

\paragraph{Density elimination} The term $\mu \brac{p_j}$ in equation \ref{Eq:cconv_primal} is introduced to eliminate the influence of input point density. 
If we remove the eliminating term, the convolution operation will then produce $\mathbf{f}^{\brac{\ell + 1}} \brac{x} = \int{\mathbf{W}\brac{\tau}^T \mathbf{f}^{\brac{\ell}} \brac{x + \tau} \mu \brac{x + \tau}} d \tau$, as it actually computes the expected value $\mathbb{E} \sbrac{\mathbf{W}\brac{\tau}^T \mathbf{f}^{\brac{\ell}} \brac{x + \tau}}$ by definition and the domain of integration has changed.
So that the term $\mu \brac{x}$ will keep being multiplied again and again through layers unless we eliminate it. 
This will not be a problem in conventional convolutional layers since $\mu \brac{x}$ represents a uniform distribution. 
Note that this treatment is still far from the best solution as it still limits the feature maps to have the same "support" as input point distribution $\mathit{P}\brac{X}$, and the estimation of $\mu \brac{x}$ from an empirical distribution is nontrivial. 
We could see the product of this convolution layer $\ell$, a point cloud with features $\mathbf{f}_{i}^{\brac{\ell}}$ as a sample from the true feature map $\mathbf{f}^{\brac{\ell}}$ using the sampling points provided in the input point cloud.
Since the feature map could also have non-zero values outside the point cloud, where we have omitted, we are still far from the best solution.
However, in practice we found that simply removing this term $\mu \brac{p_j}$ still works fine, therefore we will do so in the following descriptions for simplicity. 
We will further discuss about this issue in section \ref{Chpt:convExt}.

\paragraph{Parameterization} We parameterize the function $\mathbf{W}_\theta^{\brac{\ell}}$ by a multi-layer perceptron (MLP) with 2 hidden layers and leaky ReLU \citep{lrelu1, lrelu2}, training jointly with other part of our model. 
Since the space $\mathbb{R}^{c^\ell \times c^{\ell + 1}}$ could be too big to compute the MLP and its gradients for training, we firstly simplified the resulting matrix by a simple decomposition $\mathbf{W} = \mathbf{A} \mathbf{\Phi}$ \footnote[3]{We omitted $\ell$ for simplicity.}, with $\mathbf{A} \in \mathbb{R}^{c^{\ell} \times d}$ and $\mathbf{\Phi} \in \mathbb{R}^{d \times c^{\ell + 1}}$ and $d \ll c^{\ell}, c^{\ell + 1}$. 
Following an intuition that $\mathbf{A}$ collects input feature information, producing $d$-dimensional proxy-features, and $\mathbf{\Phi}$ collects spatial information, we therefore use $c^{\ell + 1}$ different $\mathbf{A}$'s for different output channels, and $\mathbf{\Phi}$ are then parameterized by MLP as $\mathbf{\Phi}_{\theta}\brac{\cdot} \in \mathbb{R}^3 \rightarrow \mathbb{R}^{d \times c^{\ell + 1}}$. 
Hence only $\mathbf{\Phi}$ is spatial-sensitive and $\mathbf{A}$ remains isotropic w.r.t. local positions. 
Finally we have:
\begin{equation} \label{Eq:final_cconv}
    \mathbf{f}_{i}^{\brac{\ell + 1}} = \mathbf{b}^{\brac{\ell}} + \sum_{j \in \mathcal{N}\brac{i}}
    {\mathbf{1}^T \brac{
        \text{reshape}\brac{\mathbf{\Lambda}^T \mathbf{f}_{j}^{\brac{\ell}}}
        \otimes
        \mathbf{\Phi}_{\theta}\brac {\brac{p_j - p_i}}
    }}
\end{equation}
where $\mathbf{\Lambda} \in \mathbb{R}^{c^{\ell} \times \brac{d c^{\ell + 1}}}$ is $c^{\ell + 1}$ of $\mathbf{A}$'s stacked in columns, $\text{reshape}\brac{\cdot}$ reshapes a vector from $d c^{\ell + 1} \times 1$ to a $d \times c^{\ell + 1}$ matrix, $\otimes$ denote the element-wise multiplication, $\mathbf{1} \in \mathbb{R}^{d}$ denotes an one-valued vector for summation. 
The trainable parameters within a layer $\ell$ are $\mathbf{b}^{\brac{\ell}} \in \mathbb{R}^{c^{\ell}}$ as the bias vector, $c^{\ell}$ of different $\mathbf{A}$'s as $\mathbf{\Lambda} \in \mathbb{R}^{c^{\ell} \times {d c^{\ell + 1}}}$ as the channel-wise weights, and $\theta$ of the spatial weight MLP $\mathbf{\Phi}_{\theta}\brac{\cdot} \in \mathbb{R}^3 \rightarrow \mathbb{R}^{d \times c^{\ell + 1}}$. 
In most of our experiments, we empirically choose $d = 2$ as a good trade-off between computational cost and model performance. 
We practically found that overall performance could be improved by using batch-renormalization \citep{BrN} inside the spatial weight MLP $\mathbf{\Phi}_{\theta}$, therefore we used them in our work.

\begin{figure}[t]
    \makebox[\textwidth][c]{\includegraphics[width=1.0\textwidth]{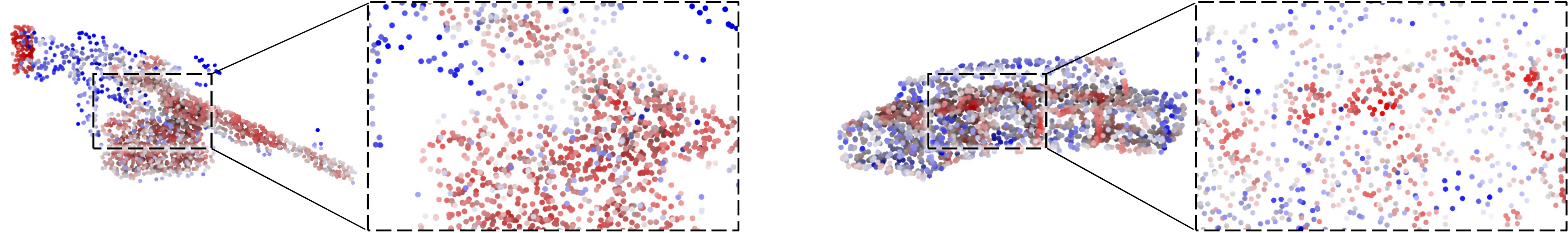}}
    \caption{Non-uniform distribution of point clouds. Densities were estimated with kernel density estimation, with the same hyper-parameters described in section \ref{Chpt:convExt}. As for the color scale, red stands for dense areas, while blue stands for sparse areas. The brightness of a point is caused by shadows calculated from screen space ambient obscurance.}
    \label{fig:pc_distribution}
\end{figure}

\subsection{Graph-based point cloud encoder} \label{Chpt:AEEnc}
As observed from figure \ref{fig:overallArch}, the encoder was composed of $K$ fine-to-coarse blocks connected with the farthest-sampling layers as pooling layers as suggested in \citep{PointNet++}. 
Each block builds a k-Nearest Neighbor Graph (k-NNG) of the input point cloud in the very beginning. 
Then the proposed convolutional layers were used in each block iteratively in order to effectively extract local features around the convolution centroid. 
After each convolution, we apply a leaky ReLU as a non-linear activation function.
We will also apply any normalization layers right before the non-linear activation function, as we will do so in experiments.

We also use additive residual connections \citep{ResNet} in our encoders to avoid vanishing gradients and accelerates the convergence of our model. 
Those connections are shown as black arrows in figure \ref{fig:overallArch}, while a connection from layer $a$ to layer $b$ is a simple addition:
\begin{equation}
    \mathbf{f}^{\brac{b}}_{res} = \mathbf{f}^{\brac{b}} + \mathbf{f}^{\brac{a}}
\end{equation}
Note that this operation is only possible when layer $a$ and layer $b$ are supported by the same point cloud with same number of points.

\paragraph{Vectors instead of clouds as the latent code} This part is optional. 
In order to obtain a latent vector, which is more widely used in representation learning literature and prior works, we simply add a global read-out layer over the final output of the encoder (the most coarse graph).
This layer is equalivent to a pooled cloud with a single node located in the origin point, and we do convolution from the last coarsened cloud to this 1-node cloud.
We also appended 2 extra fully-connected layers after the 1-node convolution to better mix the features.

\subsection{Conditional decoder} \label{Chpt:AEDec}
As we have informative-rich latent point cloud as a latent code representation, it is straightforward to let the final product of decoding (reconstruction) to be a mixture of the decoding results of every latent point, within the local space of that point. 
In precise, as indicated in figure \ref{fig:overallArch}, we consider a uniform mixture model and use a conditional generator to populate points around a given latent point.
Inspired by \citep{StyleGAN}, we use Adaptive Instance Normalization (AdaIN) \citep{AdaIN} to condition the generator. 
The AdaIN operation is defined as:
\begin{equation}
    {\text{AdaIN}}\brac{\mathbf{X}, \mathbf{y}_{\mu}, \mathbf{y}_{\sigma}} = \mathbf{y}_{\sigma} \frac{\mathbf{X} - \mathbf{\mu} \brac{\mathbf{X}}}{\mathbf{\sigma} \brac{\mathbf{X}}} + \mathbf{y}_{\mu}
\end{equation}
Where $\mathbf{X} \in \mathbb{R}^{N \times c}$ is the feature matrix with $N$ points and $c$ feature channels, $\mathbf{y}_{\mu}, \mathbf{y}_{\sigma} \in \mathbb{R}^{c}$ are vectors used for conditioning, $\mathbf{\mu} \brac{\mathbf{X}}, \mathbf{\sigma} \brac{\mathbf{X}} \in \mathbb{R}^{c}$ are the mean and standard deviation for each channels of $\mathbf{X}$, respectively. 
This operation first normalizes $\mathbf{X}$, and then apply a translation and scaling by $\mathbf{y}_{\mu}, \mathbf{y}_{\sigma}$ as the conditioning step.

As illustrated in figure \ref{fig:overallArch}, our decoder was composed by a point-wise MLP (with leaky ReLU) with 5 hidden layers, while taking a uniformly distributed random noise as an input. 
AdaIN is applied to each decoding layers while $\mathbf{y}_{\mu}, \mathbf{y}_{\sigma}$ are calculated layer-dependently via different networks, transforming corresponding latent point feature to $\mathbf{y}_{\mu}$ and $\mathbf{y}_{\sigma}$. 
The decoding network now learns a transformation that maps a trivial distribution to the desired distribution, while conditioning on latent features.
We observed a significant performance boost by applying AdaIN as the conditioner instead of concatenation, e.g. $\mathbf{y} = \mathbf{\Phi}_{\theta}\brac{\text{concat} \brac{\mathbf{z}, \mathbf{l}}}$ where $\mathbf{z}, \mathbf{l}$ are random noise vector and latent features respectively.

After we get the local point cloud of each latent point, we simply translate them to the global space (without rotation) and take the union of them as the final reconstruction, which would be used to calculate a distance with the ground-truth point cloud as a learning objective.

\subsection{Training the Auto-Encoder} \label{Chpt:AETrain}

We used the \emph{ADAM} \citep{Adam} optimizer to minimize a statistical-distance-based loss function between reconstructed clouds and the reference cloud. 
All parameters within our auto-encoder model were trained jointly via back-propagation.

\subsubsection{Loss functions} 

As point clouds are point distributions, statistical distances become a rational choice for the loss function, or the objective function, that we want to minimize between the ground-truth cloud and the reconstructed cloud. 
As our decoder is a conditional generator, the loss function acts as a non-parametric discriminator here as in GAN literature. 
Common statistical distance such as the Kullback–Leibler divergence, Total variation distance, etc. are nontrivial to compute for empirical distributions, though there are prior works aims to estimate such distance for empirical distributions \citep{KLEP}.
Some of them (e.g. Kullback–Leibler divergence) are known as they do not provide gradients (constant) or even no proper definition on non-overlay part of the supports of two distributions \citep{WGAN_pre}. 
Therefore we consider following statistical distances as they have been studied for point clouds in prior works \citep{FHQ, fold, rawgenpc}.

\paragraph{Notations} We denote a $N$ dimensional point cloud as a set of points $S = {x : x \in \mathbb{R}^N}$ which follows a $N$ dimensional distribution $P \brac{\mathit{X}}$ with PDF $\mu \brac{x}$. We also denote the euclidean distance between point $x \in S$ and point $y \in S$ as $\norm{x - y}$, and the reference and candidate point clouds $S_g$ and $S_r$ respectively.

\begin{figure}[t]
    \makebox[\textwidth][c]{\includegraphics[width=1.0\textwidth]{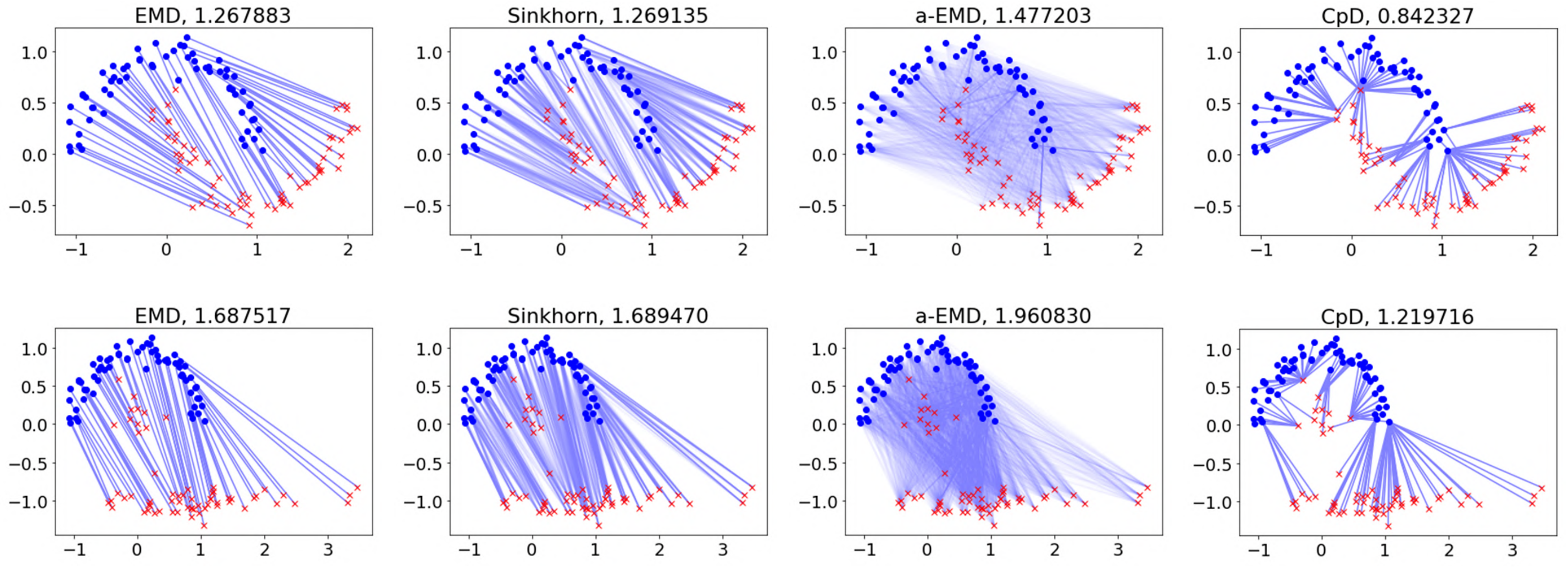}}
    \caption{Different distance metric under toy datasets (EMD: exact Earth Mover's distance, Sinkhorn: Sinkhorn distance, a-EMD: aunction based EMD, CpD: Chamfer pseudo-distance). Blue lines indicate the weight in the transport matrix used for over-all distance calculation. Note how Chamfer pseudo-distance failed to capture a consistent distance between distributions by assign lots of points in one distribution to only one point in another distribution. Meanwhile, although Sinkhorn distance seems better than auction-based EMD (a-EMD) in this figure, as it is more similar to exact EMD, it is relatively hard to use such distance in practice due to the numerical obstacles of Sinkhorn iteration when the number of points becomes bigger.}
    \label{fig:distances}
\end{figure}

\paragraph{Chamfer pseudo-distance} It used to be a popular choice as it is easy to compute in parallel and works well in practice. 
Chamfer pseudo-distance\footnote{Since the triangle inequality does not hold for Chamfer pseudo-distance, we used the term "pseudo".} is defined as: 
\begin{equation}
    d_{\text{CD}} \brac{S_r, S_g} = \sum_{x \in S_r}{\min_{y \in S_g}{\norm{x - y}}} + \sum_{y \in S_g}{\min_{x \in S_r}{\norm{x - y}}}
\end{equation}
However, as \citep{rawgenpc} points out, Chamfer pseudo-distances are "blind" to the density of two distributions, often resulting a non-uniform reconstruction result. 
As illustrated in figure \ref{fig:distances}, some well-placed reconstructed points could "block" the loss function of a point cluster in ground-truth, even if they have a dramatic difference in quantities. 
Thus, we do not use Chamfer pseudo-distance as our loss function.

\paragraph{Earth Mover's distance and Optimal Transport} Earth Mover's Distance (EMD) is an ideal choice for point cloud auto-encoders as it assumes a strict "1-to-1 mapping" of mass between source and target distributions. 
EMD is defined as (a special case of Wasserstein distance):
\begin{equation}
    d_{\text{EM}} \brac{\mu_r, \mu_g} = \inf_{\gamma \in \Gamma \brac{\mu_r, \mu_g}}{\int_{\mathbb{R}^N \times \mathbb{R}^N}{\norm{x - y} d \gamma \brac{x, y}}}
\end{equation}
Where $\Gamma \brac{\mu_r, \mu_g}$ is the set of all possible joint distributions with marginals $\mu_r$ and $\mu_g$.

Or as in its discrete form (with a slight abuse of notation):
\begin{equation} \label{Eq:OT}
\renewcommand\arraystretch{1.3}
    \begin{array}{rrl}
        & d_{\text{EM}} \brac{S_r, S_g} =&\min_{\gamma}{\inp{\gamma}{\mathbf{M}}_F}\\ [2pt]
\text{where}& \mathbf{M}_{xy} =&\norm{x - y}, \qquad x \in S_r \quad y \in S_g \\ [5pt]
\text{s.t.} & \gamma 1 =&r \\ [2pt]
            & \gamma^T 1 =&g \\ [2pt]
            & \gamma \geq&0
    \end{array}
\end{equation}
Where $\abs{S_r} = \abs{S_g} = m$, while $r, g \in \mathbb{R}^m, r_i = g_i = \frac{1}{m}$ is an uniform vector represents sample weights in both point clouds, and $\inp{\cdot}{\cdot}_F$ denotes the Frobenius inner product of matrices. 
$\gamma$ is a doubly stochastic matrix, represents the transport plan between cloud $S_r$ and cloud $S_g$ as it should satisfy the constraints $\gamma 1 = r, \gamma^T 1 = g$ (i.e. plan an exact match of mass from source to destination). 
One big drawback of EMD is they need a relatively long time to compute than Chamfer pseudo-distance and hard to parallelize. 
Fortunately, EMD is well studied by prior works of Optimal Transport (OT) and there exist many fast approximations with controllable error-rate for EMD. 
We briefly state two of them as follows \citep{EMDUB, Sinkhorn}:

The first one \citep{EMDUB} was an auction-based approach with good parallelizability, as this approach was widely used in prior works \citep{FHQ, rawgenpc, PCGAN} of learning based point cloud processing. 
We refer to this approach as auction-EMD or a-EMD in the following descriptions.
We used an implementation from \citep{FHQ} in our work. 
As stated in \citep{FHQ}, this approach gives highly accurate results with approximation error on the magnitude of 1\% with 3-D point clouds and 1024 points each.

The second approach, namely the Sinkhorn distance \citep{Sinkhorn} was developed recently in OT. 
It is done by using an additional entropy-based regularization term on the original OT problem (equation \ref{Eq:OT}):
\begin{equation} \label{Eq:Sinkhorn}
    \begin{array}{c}
         \min_{\gamma}{\inp{\gamma}{\mathbf{M}_F} - \lambda h \brac{\gamma}} \\ [5pt]
         \text{s.t.} \gamma 1 = r ,\quad \gamma^T 1 = g ,\quad \gamma \geq 0
    \end{array}
\end{equation}
Where $h \brac{\gamma} = - \sum_{i, j}{\gamma_{ij} \log{\gamma_{ij}}}$ is the entropy of $\gamma$, as $\gamma$ is a doubly stochastic matrix. 
According to \citep{Sinkhorn}, adding an entropy regularization term enforces a simple structure on the optimal regularized transport matrix $\gamma$, which could be solved iteratively using Sinkhorn iteration. 
We refer the readers to \citep{Sinkhorn} for further details. 
This approach is also easy for parallelization with great accuracy. 
Sinkhorn distance requires a great numerical capability to compute accurately, which could be time-consuming and hard for implementation.
We have implemented it using TensorFlow \citep{tf}, with different numerical capabilities, namely FP32 for single-precision floating-point numbers \emph{float32} and FP64 for double-precision.
We observed a performance boost while using precise FP64-based calculation of Sinkhorn distance as the loss function compared to a-EMD.

\subsection{Extension to the point cloud convolution kernel} \label{Chpt:convExt}

As we have shown in section \ref{Chpt:conv}, feature maps in each layer were influenced by the distribution of point clouds. 
A simple approach is already shown in section \ref{Chpt:conv}, as we refer to density elimination. 
We use kernel density estimation methods (truncated to the 1-neighborhood within the k-NNG) on each point to estimate the density $\mu \brac{x}$. 
Radial basis function (RBF) kernels $K\brac{x, x'} = \exp{\brac{- \frac{\norm{x - x'}^2}{2 \sigma^2}}}$ are a common choice for such task, and we practically choose kernel size $\sigma = \mathbb{E}_{x, y \in S}\sbrac{\norm{x - y}}$ as given point cloud $S$ and $x, y$ are distinct neighbors in the k-NNG. 
We apply the density elimination by divide the feature maps (as point cloud with features) $\mathbf{f}_{i}^{\brac{\ell}}$ in each intermediate layer $\ell \geq 1$ by the estimated input point cloud density.

Being inspired by the widely used max-pooling operation for combining features in point clouds \citep{PointNet, deepsets, fold, PointNet++, rawgenpc}, we also constract another well-working alternative approach, by substituting summation with maximum in equation \ref{Eq:final_cconv}:
\begin{equation} \label{Eq:final_cconv_max}
    \mathbf{f}_{i}^{\brac{\ell + 1}} = \mathbf{b}^{\brac{\ell}} + \max_{j \in \mathcal{N}\brac{i}}
    {\mathbf{1}^T \brac{
        \text{reshape}\brac{\mathbf{\Lambda}^T \mathbf{f}_{j}^{\brac{\ell}}}
        \otimes
        \mathbf{\Phi}_{\theta}\brac {\brac{p_j - p_i}}
    }}
\end{equation}
Where the maximum is applied channel-wise. 
We call this variation a "max-evidence" process, as we intuitively state as follows. 
Firstly, each point contributes to some evidence, where the confidence of that evidence becomes higher as the point stays in better positions. 
Then a max-pooling over all neighbors was applied to collect all evidence without letting evidence from different points overlap, that is, different points are not allowed to contribute to the same evidence multiple times.
Finally, by a linear transformation, we collect all the evidence to calculate the features, as features could be supported by multiple pieces of evidence. 

The above 2 processes significantly benefits the model for auto-encoding tasks in our experiments. 

However, as we stated in section \ref{Chpt:conv}, we are still limited by the constriction on the "support" of feature maps, where they are forced to have the same "support" with the input point cloud.
Perhaps, a resampling step could solve this problem, as in \citep{SplatNet} which resamples the feature maps to a regular hexagonal lattice. 
However a regular lattice may not be the optimal solution, as it can neither extend to higher-dimensional cases, e.g. "point cloud" where each point is actually an $H \times W \times 3$-D image instead of 3-D points, nor accurate since resampling everything to regular grids could lose more information than an optimal resampling method. 
As solving this problem will be a far step beyond our current work, we leave it as further research directions.

\section{Latent space simulation: A proof of concept}
\subsection{Interaction Networks} \label{Chpt:IN}
We adopted the structure of Interaction Network (IN) \citep{IN} as a learnable latent space dynamics estimator. 
IN is a graph network based on the message-passing principal \citep{MPNN}. 
For a given relation graph with vertices $o_i^{\ell}$ and edges $e_i^{\ell}$ between vertices $u_i^\ell$ and $v_i^\ell$, all in time step $\ell$, we could briefly state the definition of an IN as follows:
\begin{equation} \label{eq:IN}
    \begin{array}{rl}
        e_i^{\ell + 1} = & \Phi_E \brac{\brac{u_i^\ell ; v_i^\ell} ; e_i^\ell} \\[4pt]
        o_i^{\ell + 1} = & \sum_{e_i \in \mathcal{N} \brac{o_i}} {\Phi_O \brac{e_i^{\ell + 1}}}
    \end{array}
\end{equation}
Where $\brac{\cdot;\cdot}$ denotes channel-wise concatenation, $\Phi_E$ and $\Phi_O$ are parametric functions which stands for edge and vertex networks, $\mathcal{N} \brac{o_i}$ denotes the incoming edges for vertex $i$. 
In practice, $\Phi_E$ and $\Phi_O$ are often implemented as per-vertex or per-edge MLPs. 
A single layer in IN is a combination of an edge-step and a vertex-step, as shown in equation \ref{eq:IN}.

\subsection{Latent space simulator} \label{Chpt:sim}
\paragraph{Model} We implemented a simple IN with 1 layer per time step, with tanh function $\tanh x = \frac{e^x - e^{-x}}{e^x + e^{-x}}$ as activation functions. 
By $n$ layer per time step, we mean that we do $n$ iterations as in equation \ref{eq:IN} within a simulation time step to get the results. 
Vertex features $o_i$ are obtained from our point cloud auto-encoder as shown in section \ref{Chpt:AE}, concatenated with global positions of latent points.
Edges are constructed as a k-NNG at each time step, and edge features $e_i^{\ell}$ are a concatenation of relative positions and distance between vertices. 
Note we discarded the edge features every step. 
After a time step, we use a single linear transformation to obtain the movement of a latent point $\Delta p \in \mathbb{R}^3$ and apply it to the point as $p_i = p_i + \Delta p_i$. We also used residual connections inside this IN as $o_i^{\ell+1} = o_i^{\ell} + \sum_{e_i \in \mathcal{N} \brac{o_i}} {\Phi_O \brac{e_i^{\ell + 1}}}$.
\paragraph{Loss function} In order to define the loss function, we first compute a transport plan $\gamma$ as in section \ref{Chpt:AETrain}, only using the positions $\in \mathbb{R}^3$. 
Then we calculate an L2 cost matrix $M$ using both position and latent features. 
Finally our loss function is defined as $L_{\text{sim}} = \inp{\gamma}{M}_F$.
\paragraph{Training} As suggested in \citep{DeepFluids}, to improve the consistency of our model during long-term simulations, we train the model iteratively for $T=50$ time steps and calculate the loss in each time step, where the only input is the initial state of the simulated system. 
We combined losses in each time step by a weighted average with exponential weights $\alpha^{\Delta t}$, where we choose $\alpha = 0.95$ in our experiments. 
We trained the simulator separately with our auto-encoder for an acceptable computation budget.

\section{Experimental Results}

\subsection{Experiment set-up} We have implemented our models and all extensions with python library TensorFlow \citep{tf}.
We used 2 different datasets in our experiments. 
For the point cloud auto-encoder, we used the ShapeNetCore dataset \citep{ShapeNetCore} as it is a large-scale dataset contains various 3-D models with annotations, and it is available in the public domain. 
ShapeNetCore \citep{ShapeNetCore} contains 51,300 unique 3-D objects under 55 different categories, such as \emph{motorbike}, \emph{table}, \emph{watercraft}, \emph{sofa}, provided as polygon meshes. 
We sampled points from mesh surfaces by CloudCompare \citep{CloudCompare}. 
We use the official training, validation and testing splits of \citep{ShapeNetCore}, as a result, there are 35,708 samples in the training set for training, 5,158 samples in the validation set for hyper-parameter tuning and 10,261 samples in the testing set for evaluation. 
We use Sinkhorn distance-based loss as the loss function by default, as if we don't specify a choice for loss functions in the following descriptions.
In particular, we choose $\lambda=0.002$ as in equation \ref{Eq:Sinkhorn}, which is only possible under double precision floating numbers (FP64) in our experiments.

We trained our model along with other models from prior works till convergence. In order to compare with the same latent representations as prior works, we also trained a latent vector version using the method we have described in section \ref{Chpt:AEEnc}. For our cloud-based model, we used the architecture as shown in figure \ref{fig:overallArch} with density-elimination (section \ref{Chpt:convExt}) and layer normalization \citep{LN}, and trained it for $30$ epochs with batch size $8$, learning rate $5 \times 10^{-5}$, $\beta_{1}$ 0.9 and $\beta_{2}$ 0.999 with ADAM optimizer. For our vector-based model, we used the same hyper-parameter as the cloud-based model, but with a global read-out convolution as described in section \ref{Chpt:AEEnc} with 2 512-d fully-connected layers appended afterward. For FoldingNet \citep{fold} we trained with a-EMD instead of Chamfer pseudo-distance with same hyper-parameters, but for $60$ epochs with batch size $16$ and without weight decay. For AE-EMD \citep{rawgenpc} we trained it for $80$ epochs with batch size $16$.

We report the mean of exact EMD (we referred to as M-EMD) between reconstructed cloud and reference cloud over the official test set, calculated via the python package \texttt{pot} \citep{pot}.

For latent space dynamics estimation as well as point cloud auto-encoders under particle dynamics scenarios, we generated two synthesis dataset, namely "FLUIDS" for fluid dynamics and "LJP" for a simple dynamics between neutral particles. 
FLUIDS was collected using mantaflow \citep{manta} using FLIP \citep{FLIP} method by a modified example scene "breaking dam" with 760 steps total, down-sampled to 380 steps, time step $0.5$ unit, grid resolution 16, 5120 particles for simulation and down-sampled to 2048 particles without re-sampling.
We modified the example scene configuration so that we could have a diverse set of initial states.
LJP was collected using our simulation program, where it only simulates the dynamics of neutral particles under Lennard Jones Potential with a numerical cutoff.
We will show more details about our dynamics dataset in the following descriptions.
We trained our latent space estimation network as described in section \ref{Chpt:sim} with $T = 50$, $\alpha = 0.95$, $7$ epochs on our entire training dataset (397K iterations), batchsize $5$ and ADAM optimizer configured as same as above.

\begin{figure}[t]
    \makebox[\textwidth][c]{\includegraphics[width=1.0\textwidth]{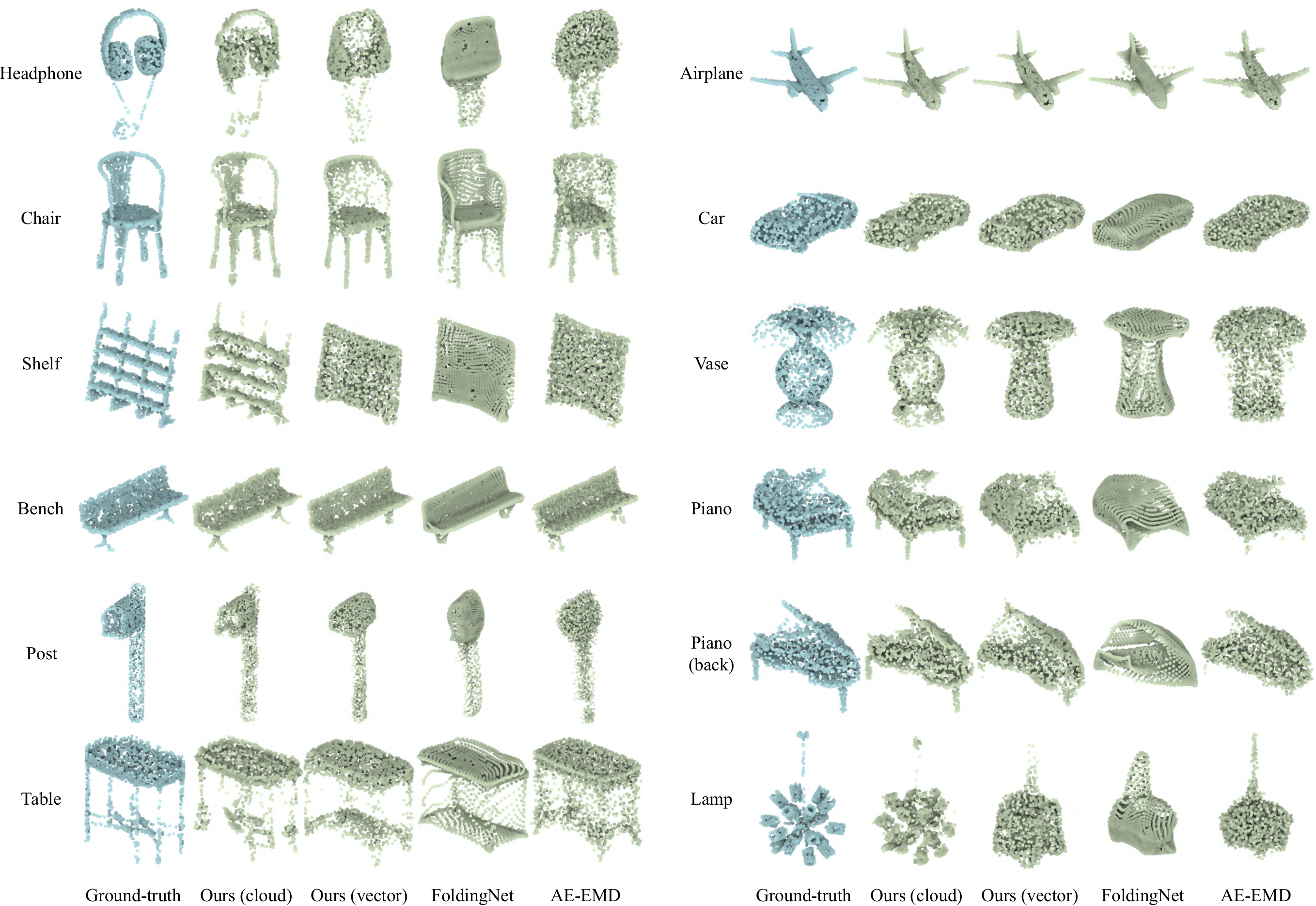}}
    \caption{Reconstruction results of our models with cloud and vector as latent code. We fairly selected some complex and uncommon objects in the ShapeNetCore \protect\citep{ShapeNetCore} testing-set for better comparison. Baseline models are FoldingNet \protect\citep{fold} and AE-EMD as in \protect\citep{rawgenpc}. Note how our latent cloud model could reconstruct complex objects such as \emph{shelf}, \emph{lamp} and \emph{piano}, and how our latent vector model could still produce good results by preserving a hollow structure for object \emph{post}, preserving the over-all, sharp structure as well as all 3 legs for object \emph{piano}, and a clean reconstruction for \emph{table}, while other baseline models struggled on those objects.}
    \label{fig:results_main}
\end{figure}

\subsection{Results and quantitative metrics}
\begin{table}[t]
\centering
\begin{tabular}{lrr}
Approach & Latent dimension & M-EMD ($\times 10^{-2}$) \\
\midrule
Random gaussian                                     &     & \texttt{17.9160} \\ [1ex]
\citeauthor{rawgenpc} (AE-EMD) \citep{rawgenpc}     & \texttt{512} & \texttt{~3.8563} \\
FoldingNet \citep{fold}                             & \texttt{512} & \texttt{~3.7528} \\ [1ex]
Ours (latent vector)                                & \texttt{512} & \texttt{~2.9290} \\
\textbf{Ours (latent cloud)}                        & \texttt{512} & \textbf{\texttt{~2.3486}} \\ [1ex]
Optimal (GT)                                        &     & \texttt{~2.1548} \\ [1ex]
\end{tabular}%
\caption{Main quantitive results. Random gaussian stands for a random gaussian with the same mean and variance compared to the training set. The M-EMD for row "Optimal (GT)" was obtained via different point-cloud samples from the same ground-truth polygon mesh, hence it states a lower-bound of error that we could ever achieve.}
\label{table:mainresult}
\end{table}
We summarized our results as shown in table \ref{table:mainresult}, as well as in figure \ref{fig:results_main}.
As stated in table \ref{table:mainresult}, our approach is significantly better than prior works, such as FoldingNet \citep{fold} or AE-EMD as described in \citep{rawgenpc}.
The improvement of our model could be easily observed from figure \ref{fig:results_main}.
We picked some complex objects in the ShapeNetCore \citep{ShapeNetCore} dataset in order to better illustrate the improvements, such as \emph{shelf} and \emph{piano}.
We also picked some uncommon objects, such as \emph{headphone} and \emph{vase}, as well as common objects such as \emph{airplane} and \emph{bench}.

Latent vector-based models such as FoldingNet \citep{fold} works better in common objects, as they could reconstruct the over-all shape of those objects, even some detailed parts.
For example, they could reconstruct the engines under the wings of a \emph{airplane}, or the supporting legs of a \emph{bench}.
As those objects appear many times among the dataset, the network is more likely to learn the shape of those objects.
And for uncommon object categories, like \emph{piano} and \emph{vase}, baseline models could only reconstruct a rough and inaccurate over-all shape of those objects.
They work even worse for the object \emph{headphone}, where baseline models failed to reconstruct nearly anything senseful.
However, our model works well in either of those objects as shown in figure \ref{fig:results_main}.

Even stylized objects under common categories are relatively hard to reconstruct, as the row \emph{chair} shows in figure \ref{fig:results_main}, baseline models failed to reconstruct that stylized chair object, and they reconstructed the special chair to a common chair.
In the same row, we could observe that our latent cloud-based model is significantly better than other models, where it reconstructs mostly of the parts in that uncommon object, including those thin armrests.
Meanwhile, the object \emph{shelf} is particularly difficult to reconstruct as it has many hollow holes inside it.
But our model still successfully reconstructed those details, and baseline models failed to reconstruct those holes.

Intuitively, we could state that our model, especially the model that used point clouds as latent representations, were strongly benefited by preserving a coarsened shape of the input point cloud in the latent space.
Our model also benefits from a layered, fine-to-coarse structure as they could handle small, non-convex details properly during the process.
Baseline models, especially \citep{rawgenpc}, mainly used a per-point MLP along with global max-pooling operation to combine features of every point, where \citep{PointNet} have shown that the final max-pooling combination operation will filter out many details, as the max-pooling roughly picks a convex hull of the original object.
FoldingNet \citep{fold} added a graph-based local max-pooling operation as a feature combination step in a local neighborhood, but it still uses a global max-pooling operation without any hierarchical process. 
So that neither FoldingNet \citep{fold} nor AE-EMD \citep{rawgenpc} will get a significantly improved performance for objects like the \emph{shelf} in figure \ref{fig:results_main}.

\subsection{Latent space dynamics}

\begin{figure}[t]
    \makebox[\textwidth][c]{\includegraphics[width=1.0\textwidth]{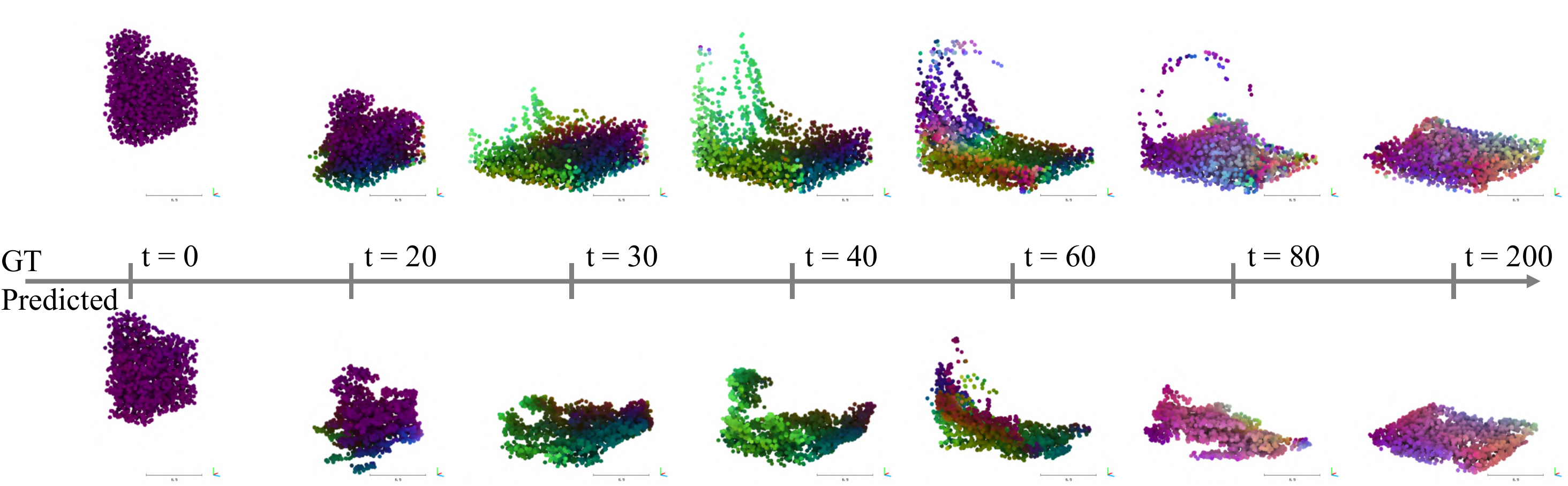}}
    \makebox[\textwidth][c]{\includegraphics[width=1.0\textwidth]{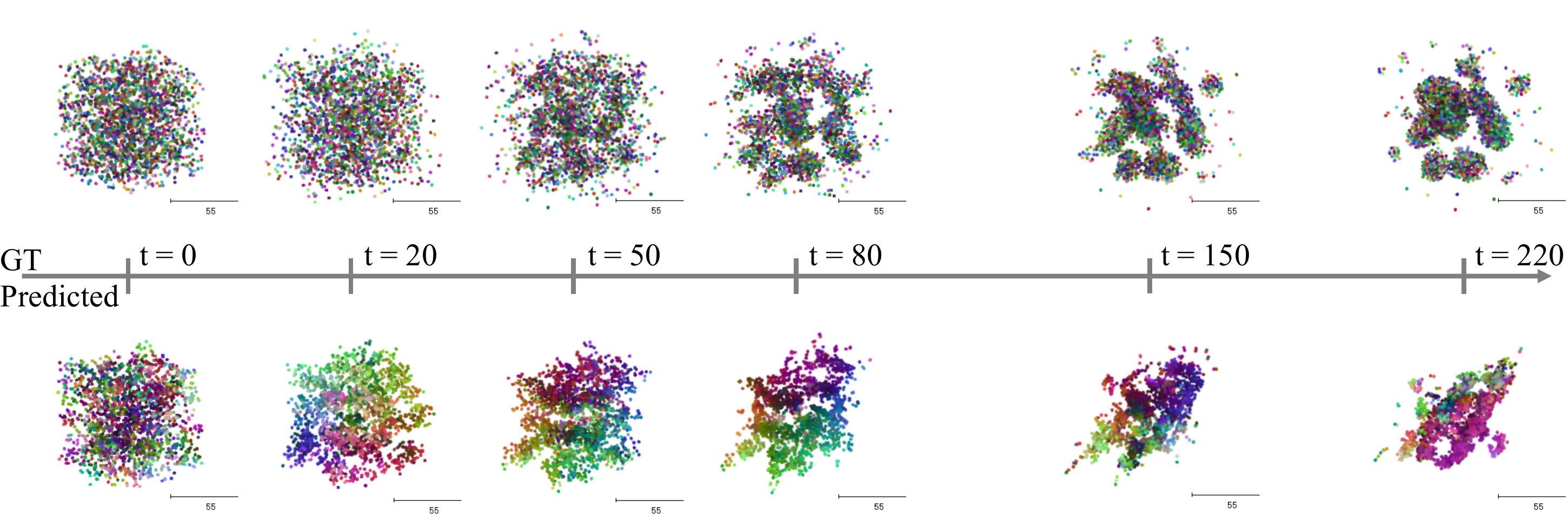}}
    \makebox[\textwidth][c]{\includegraphics[width=0.6\textwidth]{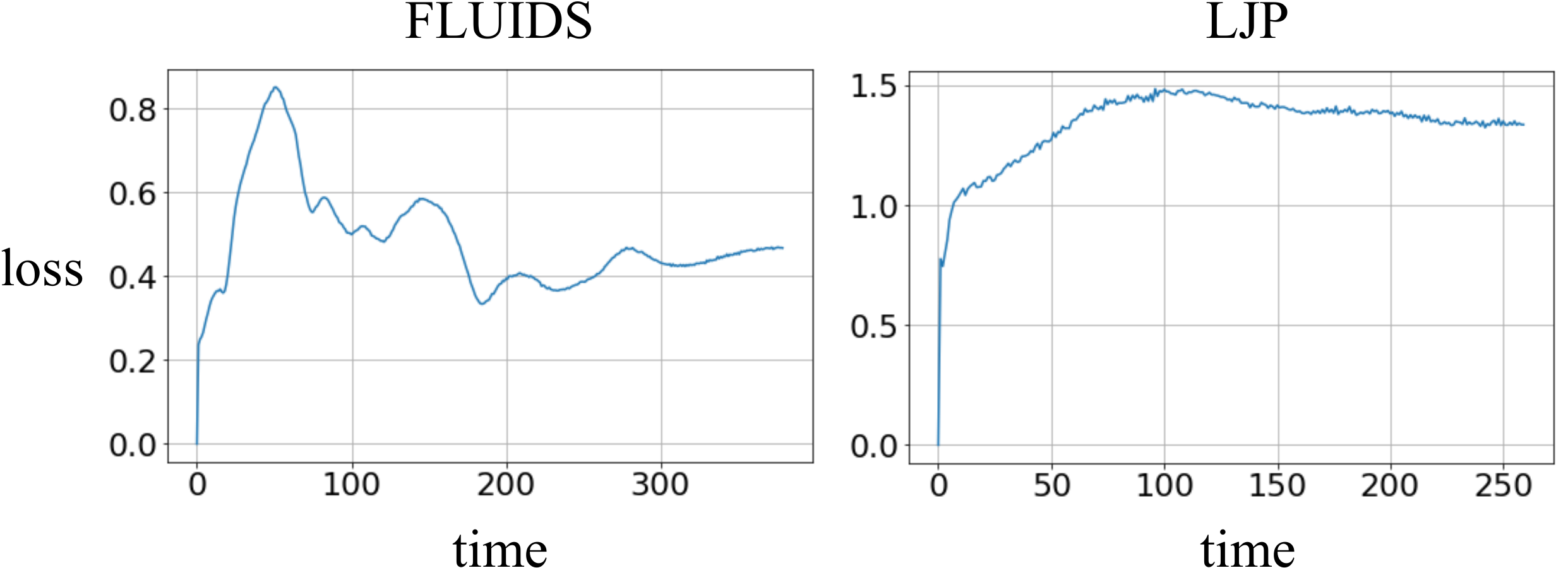}}
    \caption{Latent space dynamics simulation. Upper figure represents FLUIDS dataset, while the figure in the middle represents LJP dataset. Particles are colored by their current velocity, as our auto-encoder for latent space simulation is designed to encode and decode both position and velocity data. Note that the only input to our system is the first frame where velocity is zero every-where, and our model iteratively predicts all 379 frames (for FLUIDS) or 259 frames (for LJP) in chronological order on its own. Our model successfully catches the dynamics under latent space in this simple experiment. The lower figure indicates the mean loss (Sinkhorn distance) w.r.t. time steps during evaluation. }
    \label{fig:lSim}
\end{figure}

We trained our latent space estimator model on top of a pre-trained auto-encoder model.
Our dataset consists of 1,000 simulations with a different initial state and randomly positioned particles.
The auto-encoder used in this experiment has a 64-point latent cloud with 64 feature channels, resulting in a 4,288-d latent space.
In this experiment, we also introduced velocity along with position to our point cloud, so that each 3-d point also carry 3-d velocity information as the input feature map with 3 channels.
Our results could be seen in figure \ref{fig:lSim}, which shows that we got senseful results in this simple experimental set-up.
Since we only used a simple, single-layer Interaction Networks model without any extensions, we may expect that better results could be achieved in the near future.

\subsection{What does the model learned? A visualization of convolution kernel and activations}

\begin{figure}[t]
    \makebox[\textwidth][c]{\includegraphics[width=1.0\textwidth]{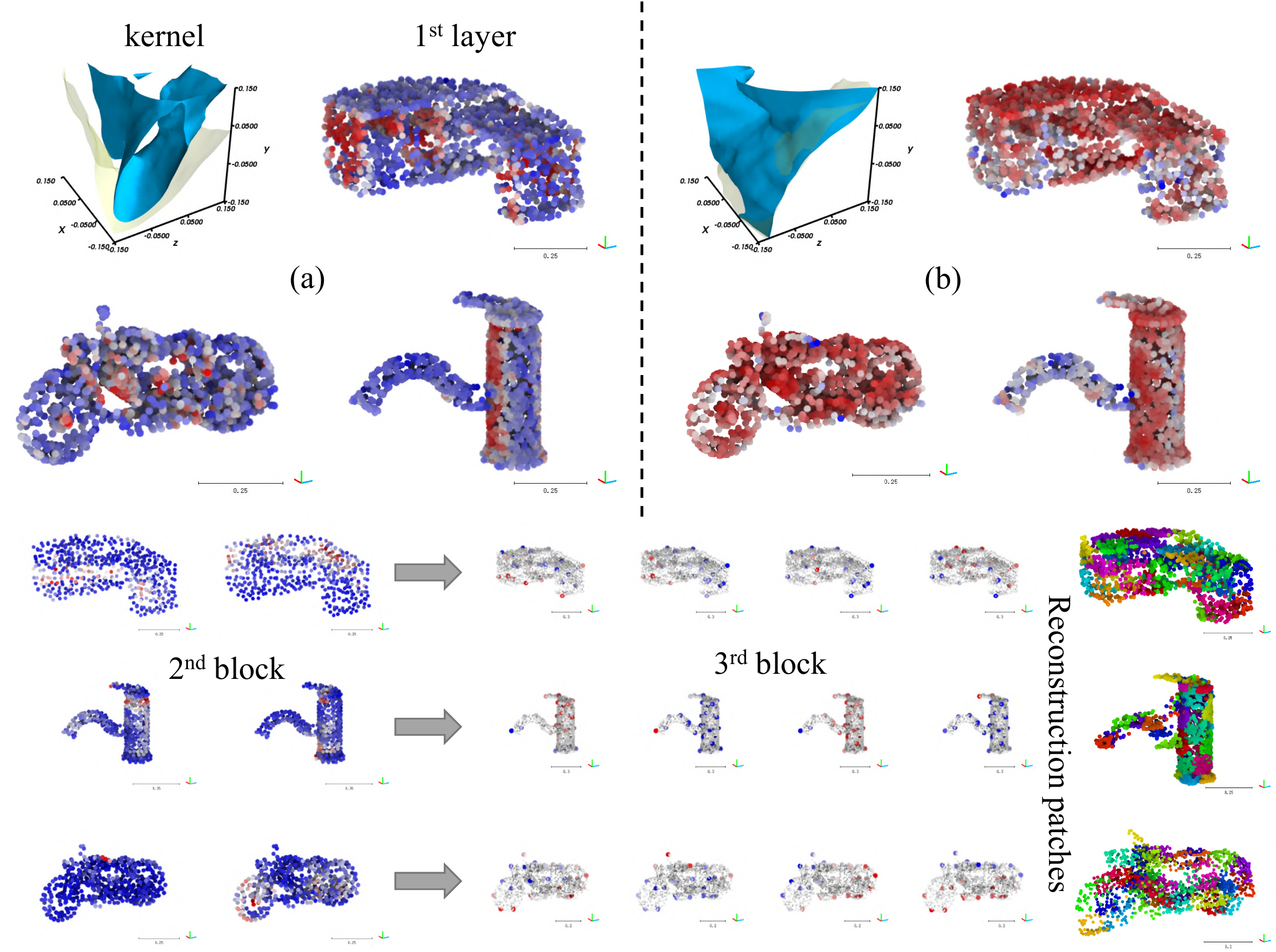}}
    \caption{kernel and activation visualizations, where we have used the same color scale as above. Note that selected channels are no longer consistent after (contains) $2^{\text{nd}}$ block, as those layers would have many channels. In addition, farthest sampling preserves a uniformly-distributed structure in the pooled clouds, which greatly benefits our model.}
    \label{fig:kernelAct}
\end{figure}

In order to obtain a better understanding of our model, we visualized the convolution kernel in each layer as well as the activations (i.e. feature maps produced) in each layer, as shown in figure \ref{fig:kernelAct}.
For convolution kernels, we rendered two isosurfaces of the entire kernel for a specific output channel (equation \ref{Eq:final_cconv}) w.r.t. position $\in \mathbb{R}^3$ and input $\mathbf{x}^{\brac{0}} = \mathbf{1}$, where light-colored isosurface locates on the 50\%th of kernel value range, and blue-solid isosurface locates on the 70\%th of kernel value range.
Only the convolution kernel of the first convolution layer was visualized, as it is the only layer that have only 1 input channel, which is convenient for visualization.
We also visualized the per-channel activation (i.e. feature maps) of each layer and colored them using the same color scale as previous sections.
We have selected the first convolution layer, the last convolution layer of each hierarchical block (without the final fully-connected layer) for visualization.

From figure \ref{fig:kernelAct} we could observe that our convolution kernel can extract local features of a given point cloud.
For example, kernel (a) catches "x-y walls" (i.e. local facelets that are orthogonal to the z-axis), which could be clearly seen in both the kernel visualization and activations on several inputs.
On the other hand, kernel (b) catches both "x-z ceilings" and "x-y walls", as seen in visualizations.
Kernel (b) is also easier to get activated compared to the kernel (a).
As the input point cloud was processed to the next hierarchical level, we could observe that some special part of it was activated, such as the handle and wheels of a \emph{motorbike}.
Then finally we could get a rich latent representation of the input cloud, where we could use it to reconstruct the cloud as a union of local patches, as figure \ref{fig:kernelAct} shown with varying colors for different patches.

\subsection{Ablation studies}
\begin{figure}[t]
    \makebox[\textwidth][c]{\includegraphics[width=1.0\textwidth]{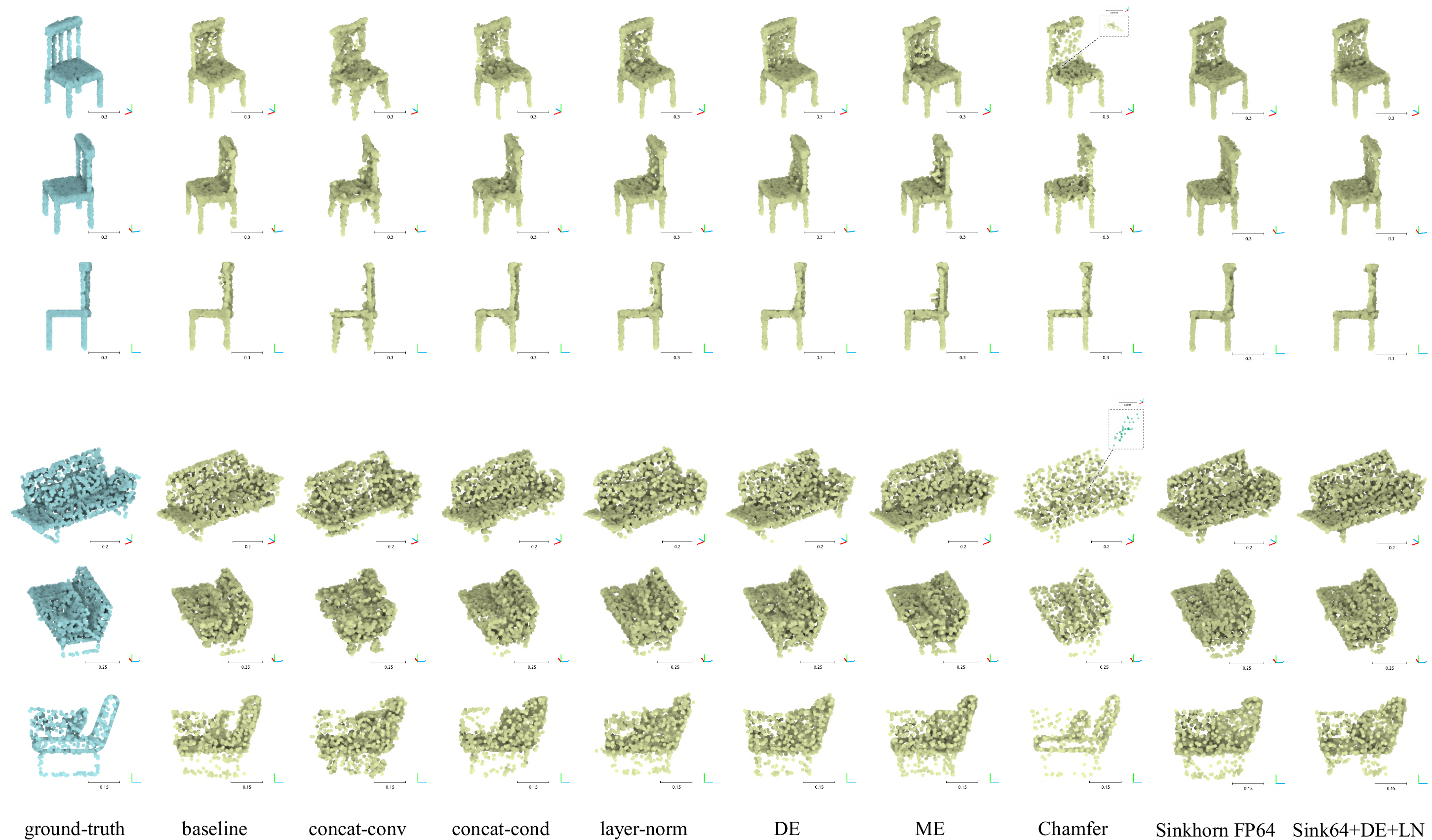}}
    \caption{Results for our ablation study, better viewed with proper magnification. Please pay attention to two small but important zoomed images for column "Chamfer", as they illustrates collapsing points and the reason of why column "Chamfer" seems to have less points than others. They actually have same number of points, but many of the points collapsed into a single "point" in column "Chamfer", probably caused by the bad properties of Chamfer pseudo-distance as we stated in figure \ref{fig:distances}. Note how concat-convolution and concat-conditioning harms the model and producing distorted reconstructions. Layer normalization, density elimination and max-evidence all regularizes the model, resulting fine and sharp results. Results for Chamfer pseudo-distance is more accurate due to an accurate point-wise distance, but it fails to preserve an uniformly distributed point cloud.}
    \label{fig:results_ablation}
\end{figure}
\begin{table}[h]
\centering
\begin{tabular}{llrr}
\multicolumn{2}{l}{Modifications} & M-EMD ($\times 10^{-2}$) & $\Delta$M-EMD ($\times 10^{-4}$) \\ 
\midrule
Baseline            &                                               & \texttt{~2.8597} &  \\ [1ex]
Random Gaussian     &                                               & \texttt{17.9160} & \texttt{+1505.63} \\ [1ex]
Convolution         & $d$ = 6                                       & \texttt{~2.7674} & \texttt{-~~~9.23} \\
                    & Concat convolution                            & \texttt{~4.1985} & \texttt{+~133.88} \\
                    & \textbf{Max-evidence (ME)}                    & \texttt{~2.6465} & \texttt{-~~21.32} \\
                    & Density-elimination (DE)                      & \texttt{~2.6651} & \texttt{-~~19.46} \\ [1ex]
Latent code         & Cloud, $16 \times 32$                         & \texttt{~2.9584} & \texttt{+~~~9.87} \\
                    & \textbf{Cloud, $\mathbf{64 \times 8}$}        & \texttt{~2.6940} & \texttt{-~~16.57} \\ [1ex]
Decoder             & Concat conditioner                            & \texttt{~3.4564} & \texttt{+~~59.67} \\ [1ex]
Normalization       & Batch renorm \citep{BrN}                      & \texttt{~2.7102} & \texttt{-~~14.95} \\
                    & \textbf{Layer norm (LN)} \citep{LN}           & \texttt{~2.6027} & \texttt{-~~25.70} \\
                    & Instance norm \citep{INorm}                   & \texttt{~2.7714} & \texttt{-~~~8.83} \\ [1ex]
Loss function       & Chamfer pseudo-dist.                          & \texttt{~6.6061} & \texttt{+~374.64} \\
                    & Sinkhorn distance FP32                        & \texttt{~2.6775} & \texttt{-~~18.22} \\
                    & \textbf{Sinkhorn distance FP64}               & \texttt{~2.6203} & \texttt{-~~23.94} \\ [1ex]
Combined            & Density-elimination + LN                      & \texttt{~2.5917} & \texttt{-~~26.80} \\
                    & \textbf{Sinkhorn FP64 + DE + LN, 18 epochs}   & \texttt{~2.3486} & \texttt{-~~51.11} \\ [1ex]
Optimal (GT)        &                                               & \texttt{~2.1548} & \texttt{-~~70.49} \\ [1ex]
\end{tabular}%
\caption{Quantitative results for our ablation studies. $d$ is as we stated in equation \ref{Eq:final_cconv}, concat convolution as $x^{\brac{\ell + 1}}_{i} = \sum_{j \in \mathcal{N} \brac{i}} {\Phi \brac{x^{\brac{\ell}}_{j} ; x^{\brac{\ell}}_{i} ; \brac{p_j - p_i}}}$, concat conditioner as in section \ref{Chpt:AEDec}, normalization as we applied to the output of each layer before activation functions, and loss functions as in section \ref{Chpt:AETrain}. }
\label{table:ablation}
\end{table}

As for the ablation study, we trained our model with different parameters and extensions, with a baseline trained as our latent cloud model stated in our main experiments, without density elimination and layer normalization, and uses a-EMD as the loss function instead of Sinkhorn-FP64.
We evaluate the model performance after substituting or adding different parts to our model as an (inverted) ablation study.
We trained all the models in our ablation experiments in a shorter time, for $10$ epochs with batch size $8$, and a larger learning rate $3 \times 10^{-4}$ and all other parameters are the same as above.

\paragraph{Latent cloud size} As observed from table \ref{table:ablation}, more latent points with fewer feature channels per point will lead to better results. 
However, since the local patches would be so small that the model may not learn useful and macroscopic representations, we kept using 32 latent points in our final model.

\paragraph{The effectiveness of our proposed convolution operation} From table \ref{table:ablation} we could find out enough evidence for the benefits of our convolution layer, compared with conventional concatenation-based convolutions. 
Meanwhile, the generalization gap between training and inference will become huge if we introduce batch-renormalization \citep{BrN} to concatenation-based convolution layers, and in contrast, it actually could benefit our convolution layers as we observed a significant boost in training and converging speed.
As normalization becomes more important as the network goes deeper, where we have used a deep model here, will lead concatenation-based convolution layers to bad results.
Nevertheless, concatenation-based convolution layers themselves are not efficient and not designed for point clouds, where our convolution layers could efficiently extract local spatial features easily.

By changing $d$ from $2$ to $6$ does improve the network performance, but only by a slight amount. 
Since the computation complexity grows nearly linearly w.r.t. $d$, this modification triples the computational cost for the convolution layers. 
Although our main computational overhead comes from the calculation of EMD and building the k-NNG and the calculation of regular matrix products in modern GPUs was well-optimized, increasing $d$ to $6$ still remains un-efficient for computation in both time and memory, compares to the slightly increased performance. 
Thus we choose $d=2$ as a sweet balance point of the trade-off shown above.

From the table, we could also observe that both density-elimination and max-evidence brings a significant improvement to the overall performance. 
As we indicated in section \ref{Chpt:conv} and \ref{Chpt:convExt}, those methods are introduced to eliminate the influence of input point cloud density. 
Since density elimination is more intuitively and theoretically supported, and it produces visually finer results compared with max-evidence as shown in figure \ref{fig:results_main} with nearly the same quantitative results, we considered using density-elimination in our final model.
However, as we stated in section \ref{Chpt:convExt}, this approach might be still far from an optimal one.

\paragraph{Choice of loss functions} As stated in table \ref{table:ablation}, Chamfer pseudo-distance behaves dramatically worse than a-EMD as well as Sinkhorn distance. 
From figure \ref{fig:results_ablation} we could observe that the results produced by Chamfer pseudo-distance mainly suffer from an unbalanced density in different parts of the reconstructed point cloud, which was already shown by \citep{rawgenpc}.

As for the Sinkhorn distance, it requires a great numerical capability to avoid numerical overflows during the Sinkhorn iteration when using lower entropy constraints, which is crucial in order to produce better results than a-EMD.
Such constraints often require the capability of hardware to efficiently compute double-precision floating-point numbers \emph{float64}, which is not efficient in older GPUs.
We found that under regular \emph{float32} settings, the lowest possible regularization term \nicefrac{1}{$\lambda$} is $1.3 \times 10 ^ {-2}$, which we refer to "Sinkhorn distance FP32", and for \emph{float64} settings the lowest possible regularization term is $2 \times 10 ^ {-3}$ as we refer to "Sinkhorn distance FP64".
Note that we first normalize the pair-wise cost matrix by dividing it element-wise by the largest entry \footnote{Since the cost matrix is non-negative, it is equivalent to the entry has the largest absolute value.} over the entire mini-batch, only for computing the transport matrix.
This could make the computation more stable and allows lower regularization terms.

Although we have above limitations, we still use Sinkhorn distance while high-speed FP64 calculation is available.
Sinkhorn distance is definitely a better choice than a-EMD in such a task, and also dramatically boosts the training process, as we stated in table \ref{table:ablation}. 
By using FP32 we got nearly the same (about 8\% longer) training time per epoch than a-EMD, and by using FP64 we are doubled the time than FP32 in modern GPUs, while 32 times slower on NVIDIA Maxwell GPUs.
Note how Sinkhorn distance-based loss leads the model to a more accurate reconstruction.

\paragraph{Conditioning methods of the decoding generator} AdaIN \citep{AdaIN} and the methods introduced in \citep{StyleGAN} is significantly better than concatenation-based methods for conditioning the decoder, as shown in table \ref{table:ablation}.
A fundamental difference of those methods is that style-based conditioner \citep{StyleGAN} fuses the conditioning vector (style) to each layer in the decoder, while concatenation-based method only appends the conditioning vector to the input of the entire decoder.
Recent developments in GANs and image synthesis \citep{StyleGAN, SAGAN, cBN, cBNGAN} also support that concatenation-based conditioning is not optimal for such tasks.

\paragraph{Normalization methods} As indicated in table \ref{table:ablation}, layer normalization \citep{LN} works best with our model, among 3 candidate methods. 
Since our model needs a relatively large memory, where only a small-sized minibatch is allowed, batch-based normalization \citep{BN} e.g. batch re-normalization \citep{BrN} still suffers from small batch-sizes.
As normalization could be crucial for deep models as they reduced the "internal covariate shift" as stated in the original work of batch normalization \citep{BN}, we choose layer normalization in our model.

\section{Future works and conclusion}
In this paper, we proposed a novel convolutional layer for irregular point clouds.
We also proposed a deep auto-encoder model based on our convolutional layer, which outperforms other baseline models and shows the benefits of our novel convolution operation, as well as a learnable latent dynamics estimator,  working well under the encoded latent space.
Our experiments have successfully shown the efficiency of our convolutional layer as well as our deep auto-encoder model.

However, as we stated in section \ref{Chpt:convExt}, we are still far from the best solution for irregular convolution operations.
Those irregular convolution operations are particularly useful since that kind of convolution could work with arbitrary data, not only constrained to 3-D geometries, but also for high-dimensional "point" distributions such as a bunch of images, videos, text, etc.
In that high-dimensional case, we could use advanced networks (e.g. convolutional neural networks, transformers \citep{AiAN}, etc.) to replace the spatial MLP $\mathbf{\Phi}_{\theta}$ as in equation \ref{Eq:final_cconv}.
This leads to an important future research direction, as efforts will be needed to extend convolution to high-dimensional, irregular points.
GANs \citep{GAN} could be useful in such cases as they provided a well-behaved statistical distance estimator in high-dimensional spaces.

On the other hand, as we proposed a novel general framework for point cloud processing, further researches may also focus on latent space simulation, point cloud compressing, classification, segmentation, etc.
As our model still has many parts to be improved (e.g. variational auto-encoders \citep{VAE}, neighbors selection \citep{PointNet++}, pooling methods), we are also interested in improving our model in those directions.

\printbibliography

\end{document}